\documentclass{article}
\usepackage{spconf,amsmath,epsfig}

\usepackage{amsmath}
\usepackage{amssymb}

\usepackage{algorithmic}

\usepackage{array}

\usepackage[english]{babel}
\usepackage{graphicx}
\usepackage{tablefootnote}

\usepackage{caption}
\usepackage{subcaption}
\usepackage{enumitem}
\usepackage{booktabs}

\usepackage[T1]{fontenc}
\usepackage{lmodern}

\usepackage{algorithm}
\usepackage{algorithmic}
\usepackage{comment}
\usepackage{color}

\usepackage{multirow}

\usepackage{amsmath}

\usepackage{array}
\newcolumntype{P}[1]{>{\centering\arraybackslash}p{#1}}
\newcolumntype{M}[1]{>{\centering\arraybackslash}m{#1}}

\makeatletter
\DeclareRobustCommand{\rvdots}{%
  \vbox{
    \baselineskip4\p@\lineskiplimit\z@
    \kern-\p@
    \hbox{.}\hbox{.}\hbox{.}
  }}
\makeatother

\usepackage{tabularx,ragged2e}

\newcolumntype{Y}{>{\RaggedRight\arraybackslash\hspace{0pt}}X}
\newcolumntype{C}{>{\Centering\arraybackslash\hspace{0pt}}X}

\graphicspath{{./}{images/}{images/binaryPlots/}}

\usepackage{tikz}
\usepackage{pgfplots}
\pgfplotsset{compat=newest}
\usetikzlibrary{shapes.geometric,arrows,fit,matrix,positioning}
\tikzset
{
    treenode/.style  = {},
    subtree/.style  = { anchor=north}
}

\title{Hashed Binary Search Tree Sampling of Dense Image Patches}

\title{Hashed Binary Search Sampling for Convolutional Network Training with Large Overhead Image Patches}

\name{Dalton Lunga, Lexie Yang, Jiangye Yuan and Budhendra Bhaduri}

\address{Computing and Computational Sciences Directorate\\Oak Ridge National Laboratory}

\begin{document}
%
\maketitle

\begin{abstract}
Very large overhead imagery associated with ground truth maps has the potential to generate billions of training image patches for machine learning algorithms. However, random sampling selection criteria often leads to redundant and noisy-image patches for model training. With minimal research efforts behind this challenge, the current status spells missed opportunities to develop supervised learning algorithms that generalize over wide geographical scenes. In addition, much of the computational cycles for large scale machine learning are poorly spent crunching through noisy and redundant image patches. We demonstrate a potential framework to address these challenges specifically, while evaluating a human settlement detection task. A novel binary search tree sampling scheme is fused with a kernel based hashing procedure that maps image patches into hash-buckets using binary codes generated from image content. The framework exploits inherent redundancy within billions of image patches to promote mostly high variance preserving samples for accelerating algorithmic training and increasing model generalization.
\end{abstract}



\section{Introduction}
Modern geographic information systems(GIS) technologies coupled with computer vision tools has played an important role in understanding social sciences related challenges. For example,  overhead imagery analysis continues to steer remote sensing applications in the direction to provide data driven decision support systems. Applications of focus include human settlement mapping,  change detection and damage assessment, infrastructure planning as well as ground object detection. These applications are benefiting from both the availability of large imagery data and increased human effort to label image pixel information through crowd sourcing applications. By leveraging these efforts, researchers are pursuing large scale automation for the long standing pixel-wise labeling challenge\cite{Paisitkriangkrai2016}. 

Among other tasks, having a large repository of labeled images benefits deep learning feature extraction and image classification. However, training inefficiencies and poor model generalization for large scale applications continue to persist as challenges that equally require investigations. We identify eliminating redundancy and noisy examples from a training sample as desirable properties to address model generalization and potential increase training efficiency in large image scene applications.

We propose a strategy for picking representative training samples to enable settlement detection over large scenes with varying terrains and snow mountains. With very high resolution, collecting training data is usually a mass selection on the screen over several neighboring pixels carrying the same information. Those selected training samples are highly redundant and can lead to severe poor model performances in particular in large convolutional networks.

Enabled by a composite kernel hashing and binary search tree sampling scheme, we demonstrate generalization efficacy through several experiments while training same convolutional network architecture, however, with different sized hash tables. More specific, we seek to investigate a novel automated strategy based on computing binary codes (hash-keys) for similar content mapping and retrieving samples with a level-order binary search tree sampling algorithm.
\section{Sampling Framework}
Within the human settlement detection task, the labels are binary, i.e. they are denoted {\em "settle"} or {\em "non-settle"} for each image patch. To preserve settlement class variability, we define an operation to map image patches onto hastable buckets using their corresponding binary codes generated by the locality sensitive kernel hashing (KLSH) technique. A binary classification based on deep convolutional network is trained with batched and sampled hash-bucket patches to give a classification label for each patch through a soft-max function. The following subsections will detail the essential components of the proposed approach.
\subsection{Dense Image Patch Hashing}
The notion of unsupervised image hashing has been explored for remote sensing applications \cite{Demir2016,Zhong2016}. We exploit it as a technique to index patches of similar content. In theory, the framework could be seen as mapping a large training database into an embedding space of image descriptors whose manifold properties can easily be interpreted via their hash bit/code representation. KLSH \cite{Kulis2009} as a technique seems to achieve just that. KLSH enables the embedding of images based on random projections $\mu_{i, i=1,2,\dots,I}$. A multivariate Gaussian distribution is assumed so that projections $\mu_{k}$  are constructed as normal random vectors approximated from sample images following the central limit theorem. The Gaussian random projections are expressed as a weighted sum of randomly selected $M$ images from the database according to:
\begin{eqnarray}
\mu_{i} =  \sum_{m=1}^{M}\omega_{i}^{(m)}\phi(\mathbf{X}_{m})
\end{eqnarray}
where $\phi(\cdot)$ is a nonlinear basis function for transforming raw images into their high dimensional kernel space representation, and $\omega_{k}$ is a coefficient vector. Kernelized locality-sensitive hashing functions 
are then constructed as: 
\begin{eqnarray}
h_{i}(\phi(\mathbf{X}_{n})) = sign(\sum_{m}^{M}\omega_{i}^{(m)}\phi(\mathbf{X}_{m})^{T}\omega_{i}^{(m)}\phi(\mathbf{X}_{n})) \\
= sign(\sum_{m}^{M}\omega_{i}^{(m)}K(\mathbf{X}_{m},\mathbf{X}_{n}) 
\end{eqnarray}
where $K(\mathbf{X}_{m},\mathbf{X}_{n}) = \phi(\mathbf{X}_{m})^{T}\omega_{i}^{(m)}\phi(\mathbf{X}_{n})$ denotes the kernel function to simplify computation of affinities in the kernel space. Following this trick, no direct computation of the basis projections $\phi(\mathbf{X}_{m})$ is needed to compute the affinities but simply values obtained from the kernel function. To proceed with hash bit calculations, for each image $\mathbf{X}_{n}$, a coefficient vector $\omega_{i}$ is estimated. Each hash bit $i,=1,2,\cdots, I$ has its corresponding $\omega_{i}$  estimated as:
\begin{eqnarray}
\omega_{i}  = K_{M}^{-\frac{1}{2}}\mathbf{e}
\end{eqnarray}
where $K_{M}^{-\frac{1}{2}}\in \mathbb{R}^{M\times M}$ is obtained from the eigen-decomposition of the kernel affinity matrix over randomly selected $M$ images. The eigen-decomposition of $K_{M}$ yields an orthogonal matrix $\mathbf{V}$ and  diagonal matrix $\mathbf{U}$ so that  $ K_{M}^{-\frac{1}{2}} = \mathbf{V}\Lambda^{-\frac{1}{2}}\mathbf{V}^{T}$. An indexing vector $\mathbf{e}$ randomly samples $t$ images from the $M$ images. Each  $\omega_{i}, i=1,2,\cdots,I$ is estimated  by randomly re-selecting $I$ times $M$ samples and re-estimating $K_{M}\in \mathbb{R}^{M\times M}$.  As such, a total of $I$ hash functions are initially generated and then applied to each image $\mathbf{X}$. This yields a hashtable with hashcodes $H_{\mathbf{X}_{n}} =[h_{1}(\phi(\mathbf{X}_{n})),h_{2}(\phi(\mathbf{X}_{n})),\cdots,h_{I}(\phi(\mathbf{X}_{n}))],n=1,\cdots, N$ for all images.
\subsection{Binary Search Tree Sampling}
\begin{figure}[!ht]
  \centering
       \includegraphics[width=.5\textwidth]{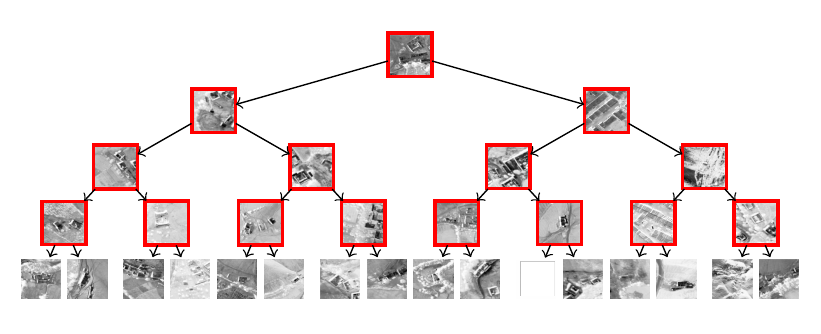}
  \caption{An illustration of a hashed binary search tree on image hashtable.  At any level $\ell$ the number of nodes is computed as $2^{\ell -1}$. Shown in \textsf{red-rectangles} is a \textsf{level-order} ($\ell=4$) sampling of the \textsf{settlement-class} image patches from a hashbucket whose binary search tree has $31$ nodes.}
  \label{fig:binary-searchtree}
\end{figure}

Binary search trees are popular extensions of the concept of linked data structures to a structure containing nodes with more than one self-referenced field. The tree is made of nodes, where each node contains a "left" reference, a "right" reference, and a data element. The topmost node in the tree is called the root. As shown in Algorithm~1, given a hashtable $\mathcal{H}$, implementing a binary search tree sampling scheme proceeds by sorting each hash-bucket based on its set of hashcodes $H_{\mathbf{X}_{n}}$. From the sorted list, the middle hashcode is designated as the root node for the tree. The sampling strategy is then instituted to prune the tree depthwise following a \textsf{level-order traversal} technique to compute the variance of sampled images. Although it would be desirable to extend our sampling technique to a multi-bucket sampling scheme, we only consider a single binary search tree for each hashbucket. Figure~\ref{fig:binary-searchtree} illustrates a hashbucket tree that is efficiently built by comparing image hashcodes to the designated root patch.  

The process is efficiently carried out offline with the variance computed as
\begin{eqnarray}
\vartheta_{\ell}  \Leftarrow \frac{\textsf{var}(\mathbf{B}_{n}^{(\ell)})}{\textsf{var}(\mathbf{B}_{n}^{(L)})}
\end{eqnarray}
where  $n$ is the bucket number whose binary tree representation is $\mathbf{B}_{n}^{(L)}$  and has height $L$.  $\mathbf{B}_{n}^{(\ell)}$ is the subtree extracted depthwise up to level $\ell$ from $\mathbf{B}_{n}^{(L)}$.
\begin{algorithm}
 \begin{algorithmic}  
 \REQUIRE $H \Leftarrow KLSH, \epsilon,  D \Leftarrow \{\}$
 \ENSURE $D$
 \WHILE{$\textsf{not at end of Hashtable} \ \mathcal{H}$}
	 	\STATE $H_{B} \Leftarrow \textsf{readNextBucket}(H)$;
	 	\STATE $H_{B} \Leftarrow \textsf{sortBucketKeys}(H_{B})$
	 	\STATE $r_n \Leftarrow \textsf{selectMidHashKey}(H_{B})$\COMMENT{pick root node};
	 	\STATE $\mathbf{B}^{(L)},L \Leftarrow \textsf{buildBinaryTree}(H_{B},r_n)$;
	 	\FOR{$\ell$=1 \TO L } 
	 			\STATE $\mathbf{B}^{(\ell)} \Leftarrow\textsf{ levelOrderTraversal}(\mathbf{B}^{(L)},\ell,r_n)$
	 			\STATE $\vartheta_{\ell}  \Leftarrow \frac{\textsf{var}(\mathbf{B}^{(\ell)})}{\textsf{var}(\mathbf{B}^{(L)})}$;
			  	\IF{$\vartheta_{\ell} \geq \epsilon$}
		            \STATE $\textrm{return}\  \mathbf{B}^{(\ell)}$
		        \ENDIF
        \ENDFOR
        \STATE $D \Leftarrow \textsf{addSubtree}(D,\mathbf{B}^{(\ell)})$;
 \ENDWHILE
 \STATE $\textrm{return}\ D$;
 \end{algorithmic}
 \caption{Binary Search Tree Sampler}
\end{algorithm}%
\subsection{Convolutional Network Architecture}
To demonstrate the efficacy of the sampling scheme, we designed a 7-layer convolutional network (CNN) based classifier through stacking of several units that perform linear or nonlinear transformation of the image patches. The units includes 4-convolutional layers, 4-maxpool layers, 2-fully connected layers, 2-dropout layers and a softmax output function. Model parameters for the convolutional filters are obtained via a stochastic gradient descent (SGD) technique based on the back-propagation framework \cite{Hinton2006,Glorot2010}. For SGD, learning rate is set to $0.00273$ via a full gridsearch hyperparameter setting, while the CNN activation is set to ReLU, filter weights initialized from a normal distribution, and the batch size set to $200$.
\section{Experiments}
Four image tiles that cover extremely varying landscapes and settlement structures are selected. A \textsf{Ground-truth} collection of $175,000$ image patches, each of size $144\times 144$ pixels, is manually cropped from the tiles. The sampling algorithm is applied on the \textsf{Ground-truth} to select $60,000$ representative patches for training a convolutional network based settlement detection model. Model evaluation is conducted using a validation set of $10800$ image patches.  For large scale generalization test, a high-resolution aerial image of size $41737 \times 28485$ covering a scene from a different province in Afghanistan is used.
\subsection{Settlement Detection Performance}
\begin{figure}[!ht]
  \centering
       \includegraphics[width=.4\textwidth]{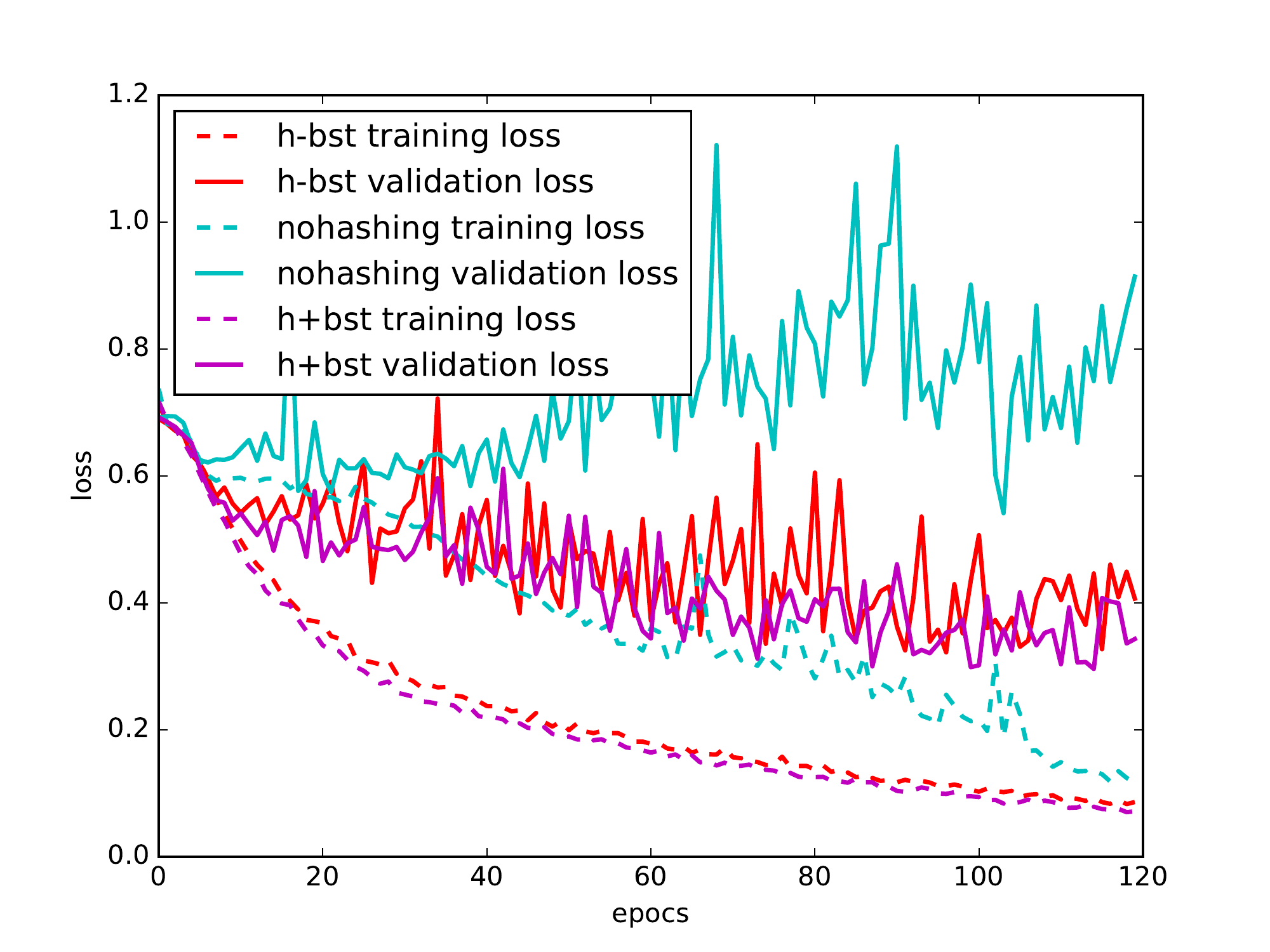}
  \caption{Training and validation $loss$ performance for different CNNs trained with three different sets of $15000$ training samples selected from \textsf{Ground-truth}.}
  \label{fig:loss-validation-curve}
\end{figure}

\begin{figure*}[t]
\begin{subfigure}{.4\linewidth}
  \centering
  \includegraphics[width=10cm,
  height=4cm,
  keepaspectratio]{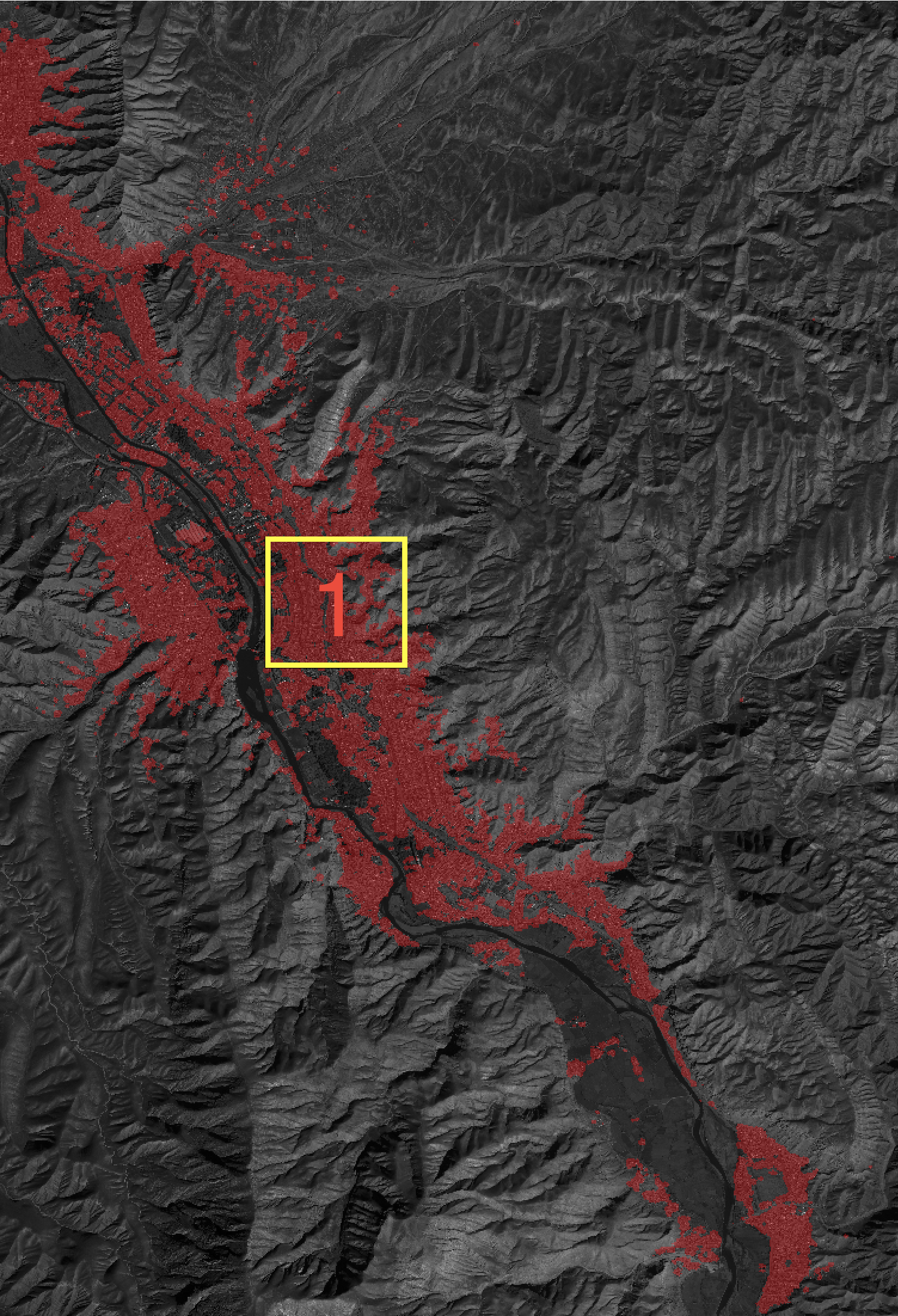}
  \caption{CNN settlement detection over Baghlan. (City over 1000sq km)}
  \label{fig:sfig1}
\end{subfigure}%
\begin{subfigure}{.6\linewidth}
  \centering
  \begin{subfigure}{.6\textwidth}
  \centering
  \includegraphics[width=6cm,
  height=4cm]{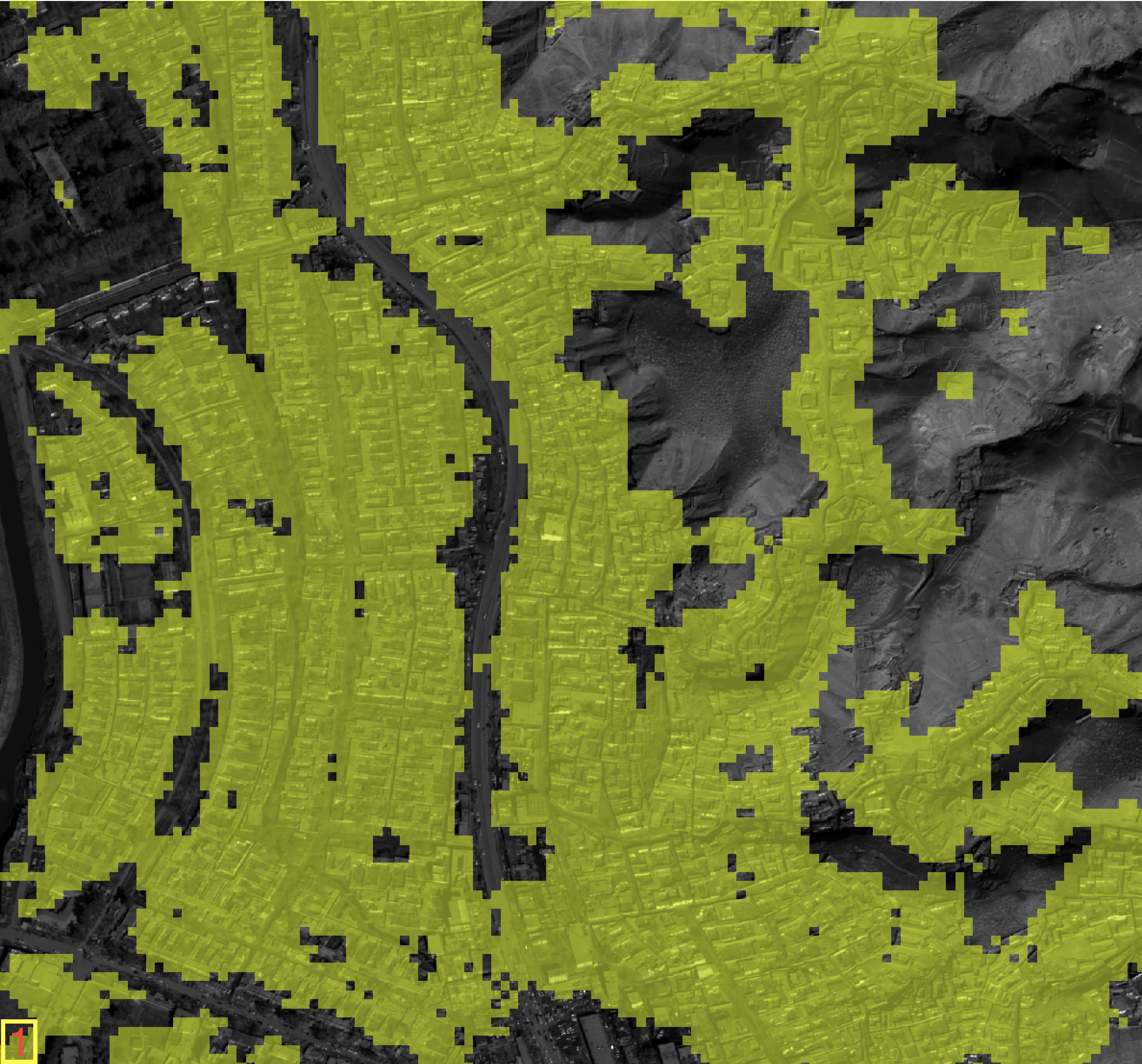}
  \caption{$60,000$ sample based output.}
  \label{fig:sfig2}
\end{subfigure}
\begin{subfigure}{.6\textwidth}
  \centering
  \includegraphics[width=6cm,
  height=4cm]{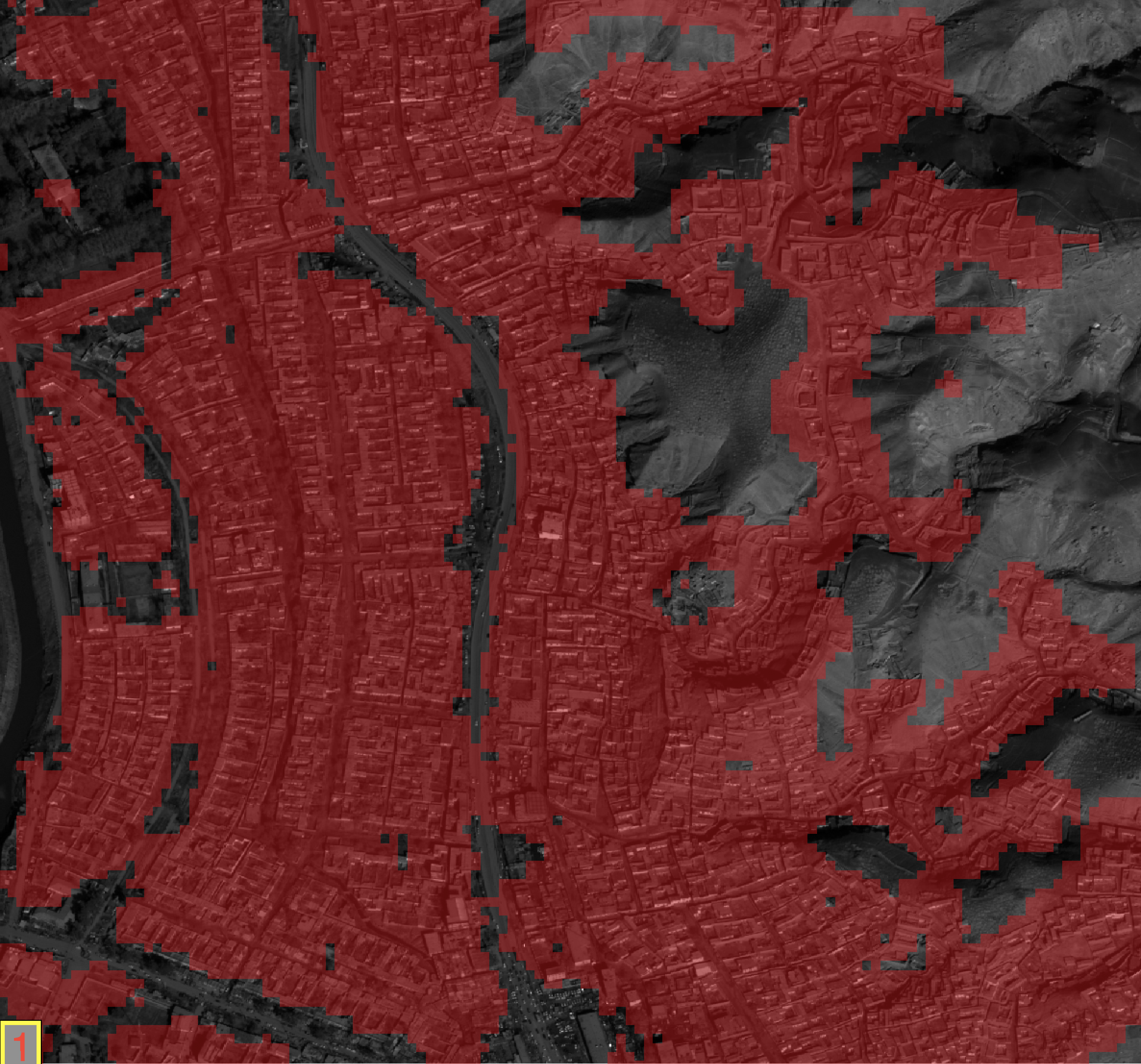}
  \caption{\textsf{Ground-truth} ($175,000$ sample) based output}
  \label{fig:sfig2}
\end{subfigure}
\end{subfigure}
\caption{Illustration of large scale CNN settlement detection output $(a)$.  In $(b)$ and  $(c)$ a close look to location marked \textsf{"1"}(from (a)) to compare CNN trained with $60,000$ samples versus a \textsf{Ground-truth} based model.}
\label{fig:settlement-detection}
\end{figure*}

Hashing and sampling performance is evaluated with three kernels including the \textsf{polynormial}, \textsf{laplacian} and the \textsf{radial basis function}. Table~\ref{tab:kernel-functions} shows classification performance results. The \textsf{radial basis  function} appears to offer better hashing thereby enabling binary search tree sampling that best represent the \textsf{Ground-truth}. Figure~\ref{fig:loss-validation-curve} also highlights the difference between selecting atmost $20$ image patches from each hash-bucket in a hashtable to make up $15,000$ training images versus sampling the same hastable for $15,000$ images using the binary search scheme of Algorithm 1. Models are defined as follows: \textsf{hashing plus binary search model(h+bst)}, and \textsf{hashing minus binary search model(h-bst)}. Both \textsf{h+bst} and \textsf{h-bst} appear less sensitive to over-fitting. However, the \textsf{no-hashing model} is observed to overfit from the $25$th epoch and exhibit slow convergence. Figure~\ref{fig:settlement-detection} demonstrates the pixel labeling and generalization capability of a \textsf{h+bst} based $60,000$ sample trained CNN on a $41737 \times 28485$ test scene. The model offers matching results when compared with the \textsf{Ground-truth} based CNN.
\begin{table}[t]
 \centering
\caption {\footnotesize Performance is reported for hashing with different kernel functions while sampling $60,000$ image patches from the \textsf{Ground-truth} collection. For reference, the \textsf{Ground-truth} based model achieves $0.949$ accuracy the "Settle" and $0.959$ on the "Non-Settle" classes.} 
\label{tab:kernel-functions} 
{\footnotesize
\begin{tabular}{cccc}
\firsthline
\hline
Kernel Function   & Metric & Settle & Non-Settle \\
\midrule
 \multirow{3}{*}{$\textsf{Radial Basis}$} & accuracy &$0.946$ & $0.956$ \\
 & precision & $\textsf{0.931}$ & $\textsf{0.942}$ \\
 & recall & $\textsf{0.964}$ & $\textsf{0.974}$ \\ 

 \multirow{3}{*}{$\textsf{Laplacian}$} & accuracy&$0.945$ & $0.954$ \\
 &   precision & $\textsf{0.928}$ & $\textsf{0.945}$ \\
 &  recall & $\textsf{0.961}$ & $\textsf{0.967}$ \\ 

 \multirow{3}{*}{$\textsf{Polynomial}$} & accuracy &$0.897$ & $0.907$ \\
 &   precision & $\textsf{0.885}$ & $\textsf{0.921}$ \\
 &  recall & $\textsf{0.89}$ & $\textsf{0.912}$ \\ 
 \bottomrule
\end{tabular} }
\end{table}
\section{Discussion}
We have presented a framework for sampling dense image patches inspired by a binary search tree with level order traversal scheme that takes advantage of the kernel space locality sensitive hashing to organize image patches. The framework demonstrates potential to select representative image patches to train large convolutional models for wide scene categorization tasks. The sampling technique demonstrated significant benefits for automated training sample abstraction over complex and large scenes. The proposed work opens up new avenues to direct research efforts toward investigating training inefficiencies, model generalization and over-fitting - all challenges that are hindering the full potential of large scale pixel labeling with deep convolutional networks.

\section*{Acknowledgement}
This manuscript has been authored by UT-Battelle, LLC
under Contract No. DE-AC05-00OR22725 with the U.S.
Department of Energy. The United States Government retains
and the publisher, by accepting the article for publication,
acknowledges that the United States Government
retains a non-exclusive, paid-up, irrevocable, world-wide
license to publish or reproduce the published form of this
manuscript, or allow others to do so, for United States
Government purposes.

The authors would like to thank Jeanette Weaver for her contribution
on selecting testing sites and preparing testing images.
\bibliographystyle{IEEEbib}
\bibliography{overheard-dcnn}

\end{document}